\documentclass[letterpaper, 10 pt, conference]{ieeeconf} 
                        
\IEEEoverridecommandlockouts                              
\usepackage{float}
\usepackage{graphicx}
\usepackage{lipsum}
\usepackage{stfloats}
\usepackage{multicol}
\usepackage{multirow}
\usepackage{amsmath,bm}
\usepackage{etoolbox}

\usepackage{array}

\usepackage{tabulary}
\usepackage{xcolor}

\usepackage{hyperref}
\hypersetup{colorlinks,linkcolor=black}
\hypersetup{colorlinks,urlcolor=black}

\usepackage[caption=false,font=footnotesize]{subfig}

\title{\LARGE \bf
DS-PASS: Detail-Sensitive Panoramic Annular Semantic Segmentation through SwaftNet for Surrounding Sensing
}

\author{Kailun Yang$^{1}$, Xinxin Hu$^{2}$, Hao Chen$^{2}$, Kaite Xiang$^{2}$, Kaiwei Wang$^{2}$ and Rainer Stiefelhagen$^{1}$
\thanks{This work has been partially funded through the project ``Research on Vision Sensor Technology Fusing Multidimensional Parameters'' (111303-I21805) by Hangzhou SurImage Technology Co., Ltd., as well as supported by Hangzhou HuanJun Technology Co., Ltd. and Hangzhou KrVision Technology Co., Ltd. (krvision.cn).}
\thanks{$^{1}$Authors are with Institute for Institute for Anthropomatics and Robotics, Karlsruhe Institute of Technology, Germany. \tt kailun.yang@kit.edu}
\thanks{$^{2}$Authors are with National Optical Instrumentation Engineering Technology Research Center, Zhejiang University, China}
}

\begin{document}

\maketitle
\thispagestyle{empty}
\pagestyle{empty}

\begin{abstract}

Semantically interpreting the traffic scene is crucial for autonomous transportation and robotics systems. However, state-of-the-art semantic segmentation pipelines are dominantly designed to work with pinhole cameras and train with narrow Field-of-View (FoV) images. In this sense, the perception capacity is severely limited to offer higher-level confidence for upstream navigation tasks. In this paper, we propose a network adaptation framework to achieve Panoramic Annular Semantic Segmentation (PASS), which allows to re-use conventional pinhole-view image datasets, enabling modern segmentation networks to comfortably adapt to panoramic images. Specifically, we adapt our proposed SwaftNet to enhance the sensitivity to details by implementing attention-based lateral connections between the detail-critical encoder layers and the context-critical decoder layers. We benchmark the performance of efficient segmenters on panoramic segmentation with our extended PASS dataset, demonstrating that the proposed real-time SwaftNet outperforms state-of-the-art efficient networks. Furthermore, we assess real-world performance when deploying the Detail-Sensitive PASS (DS-PASS) system on a mobile robot and an instrumented vehicle, as well as the benefit of panoramic semantics for visual odometry, showing the robustness and potential to support diverse navigational applications.

\end{abstract}

\section{INTRODUCTION}

Semantically interpreting the traffic scene is one of the crucial perception components for system-critical autonomous transportation and robotics applications. Currently, semantic image segmentation, which unifies detection and classification at the pixel level~\cite{yang2018unifying}, has demonstrated accurate parsing capability in standard conditions.

\begin{figure}[t]
\setlength{\abovecaptionskip}{0pt}
\setlength{\belowcaptionskip}{0pt}
\centering
\includegraphics[width=0.48\textwidth]{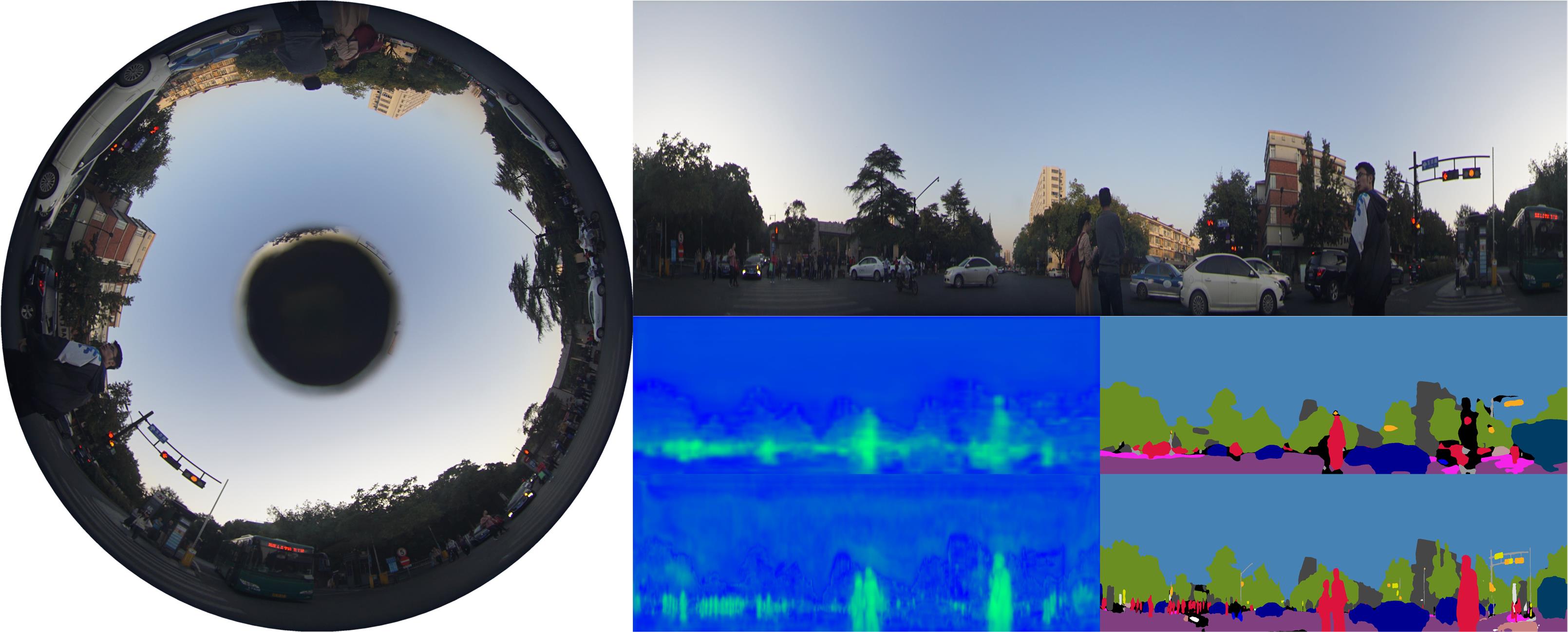}
\caption{Panoramic annular semantic segmentation. On the left: raw annular image; First row on the right: unfolded panorama; Second row: panoramic segmentation of the baseline method~\cite{yang2019can}, the classification heatmap of pedestrian is blurry; Third row: detail-sensitive panoramic segmentation of the proposed method, the heatmap and semantic map are detail-preserved.}
\label{figurelabel_concept}
\vskip-4ex
\end{figure}

However, modern segmentation algorithms pertain to work with narrow Field-of-View (FoV) pinhole cameras, although panoramic camera is gradually becoming more attractive to be integrated on Intelligent Vehicles (IV) systems. One of the essential reasons lies in that most publicly available segmentation datasets~\cite{cordts2016cityscapes}\cite{neuhold2017mapillary} contain merely pinhole images whose FoV is not as wide as panoramas. When taking a pre-trained segmentation network from pinhole imagery to the omnidirectional imagery, the performance drops significantly and often catastrophically due to the divergence of global semantic and structural information across the image domains~\cite{yang2019can}. In addition, for panoramic images, semantic segmentation is required to perform with megapixel resolution to cover the FoV as wide as 360$^{\circ}$, and perceive tiny scene elements such as pedestrians in the distance (see Fig.~\ref{figurelabel_concept}). Nevertheless, encoder-decoder networks such as~\cite{paszke2016enet}\cite{romera2019bridging} largely sacrifice the detail-sensitivity in the down-sampling process with slim up-sampling path to maintain efficiency. It is even more serious for large panoramic images, as the resolution difference is more significant between the feature map at the semantically-rich deep layer and the original panorama.

To address these gaps, we propose a network adaptation framework, which allows to re-use knowledge learned from conventional pinhole image datasets and avoid blind spots at the panorama borders, such that contemporary semantic segmenters can adapt seamlessly to panoramic images. Specifically, we put forward a real-time semantic segmentation network named SwaftNet by using attention-based lateral connections, further enhancing the capacity to exploit low-level detail-sensitive features for high-level context-critical surrounding sensing.

To verify the effectiveness of our network adaptation strategy and the detail-sensitivity of attention-augmented SwaftNet, we extend the PASS dataset~\cite{yang2019can}\cite{yang2019pass} by finely annotating more detail-critical classes such as sidewalks and pedestrians. With this extended panoramic dataset, we benchmark the performance of real-time scene parsing networks on the challenging task: panoramic annular semantic segmentation. The comprehensive set of experiments demonstrates that the network adaptation strategy is consistently positive for all state-of-the-art efficient segmenters. Moreover, the proposed attention-connected SwaftNet reaches the best performance while maintaining a fast inference speed far above the real-time constraint. 

Furthermore, the panoramic segmentation pipeline is deployed on a mobile robot and an instrumented vehicle, fully validating the robustness against real-world unseen domains and the potential to support diverse autonomous transportation applications. The framework is additionally combined with a modern panoramic visual odometry~\cite{chen2019palvo}, throwing hints on how navigation tasks benefit from the surrounding scene parsing results. The extended PASS dataset and panoramic image sequences collected by diverse systems, as well as the codes of our framework, are made publicly available at~\footnote{Datasets and codes of DS-PASS:~\url{https://github.com/elnino9ykl/DS-PASS}}. 

\section{Related Work}

\subsection{Semantic Segmentation Networks}

Semantic segmentation has experienced rapid progress since Fully Convolutional Networks (FCNs)~\cite{long2015fully}, along with U-Net~\cite{ronneberger2015u} and SegNet~\cite{badrinarayanan2017segnet} which were recognized as the milestone works. Subsequently proposed networks like PSPNet~\cite{zhao2017pyramid} and ACNet~\cite{hu2019acnet} have reached higher accuracy on diverse semantic segmentation benchmarks. However, these networks rely on heavy backbones and sophisticated structures, which significantly burden the function complexity, leaving them irrelevant with real-time applications. Following SegNet~\cite{badrinarayanan2017segnet} that has demonstrated near real-time speed
ENet~\cite{paszke2016enet} was presented as the first work to implement real-time semantic segmentation with many inspiring practical techniques. Unfortunately, ENet largely sacrifices the accuracy, unable to yield finely-grained segmentation maps.

To address this problem, ERFNet/ERF-PSPNet~\cite{yang2018unifying}\cite{romera2019bridging}, LinkNet~\cite{chaurasia2017linknet} and SwiftNet~\cite{orsic2019defense}\cite{xiang2019comparative} were proposed to strike a better balance between efficiency and accuracy. Most of these networks achieve dense predictions based on an encoder-decoder architecture starting from gradual down-sampling of the input image, then the contextually-rich feature maps are restored to the original resolution through the corresponding up-sampling process. Lateral skip connections were naturally used to merge the features from the deep layers and the spatially-rich shallow layers to be aware of detailed variance within local regions~\cite{ronneberger2015u}\cite{chaurasia2017linknet}\cite{orsic2019defense}. For panoramic images, semantic segmentation is required to perform with very large resolution to cover the wide FoV up to 360$^{\circ}$, and perceive tiny scene elements such as pedestrians at distances of over hundreds of meters in order to support safe autonomous transportation applications. Nevertheless, the resolution difference is even more significant between the large panorama and the feature map at the semantically-rich deep layer of a typical segmentation network. In this sense, with the ultimate goal of eliminating blind spots for surrounding sensing, the detail-sensitivity of a segmentation network should be further enhanced to better exploit spatial variance while maintaining the high efficiency.

\subsection{Panoramic Segmentation Approaches}

To realize larger Field of View (FoV) of semantic perception, research efforts have been made on fisheye image segmentation~\cite{deng2017cnn}, surround-view segmentation~\cite{deng2019restricted}\cite{yahiaoui2019fisheyemodnet}\cite{iordache2019low}, panoramic semantic change detection~\cite{sakurada2018weakly}, cross-view semantic segmentation~\cite{pan2019cross}, spherical semantic segmentation~\cite{zhang2019orientation} and panoramic lane marking segmentation~\cite{jiang2019dfnet}. However, most of them need several cameras to form the 360$^{\circ}$~\cite{deng2019restricted}\cite{yahiaoui2019fisheyemodnet}\cite{pan2019cross}\cite{jiang2019dfnet}. For fisheye images, the significant distortion is clearly not beneficial for upper-level tasks~\cite{deng2017cnn}, which needs to be rectified for accurate perception with respect to spatial configurations of the surroundings. For spherical input, it is required to be converted to an unfolded mesh to provide a holistic labeling of the scene~\cite{zhang2019orientation}. More importantly, deploying several cameras incurs large processing latency and a cluster of hard tasks such as calibration, synchronization and sensor fusion, which again are not advantageous for autonomous navigation applications.

For this reason, our previous work~\cite{yang2019can} introduced the task of Panoramic Annular Semantic Segmentation (PASS) to address surrounding sensing in a unified way, which only involves a single camera to capture distortion-controlled annular images, running with a single forward pass of ERF-PSPNet that was structured using a combination of ERFNet~\cite{romera2019bridging}\cite{romera2018erfnet} and PSPNet~\cite{zhao2017pyramid}. Nevertheless, the chosen network architecture is not ideally suitable for panoramic images. In spite of the decent generalizability, the parsing results are too smooth/blurry, leaving many scene details not perceived. In this work, we aim to adapt in an optimal way, making panoramic semantic segmentation not only robust but also detail-sensitive.

\section{Methodology}

\subsection{Network Adaptation Framework}

\begin{figure}[t]
\setlength{\abovecaptionskip}{0pt}
\setlength{\belowcaptionskip}{0pt}
\centering
\includegraphics[width=0.48\textwidth]{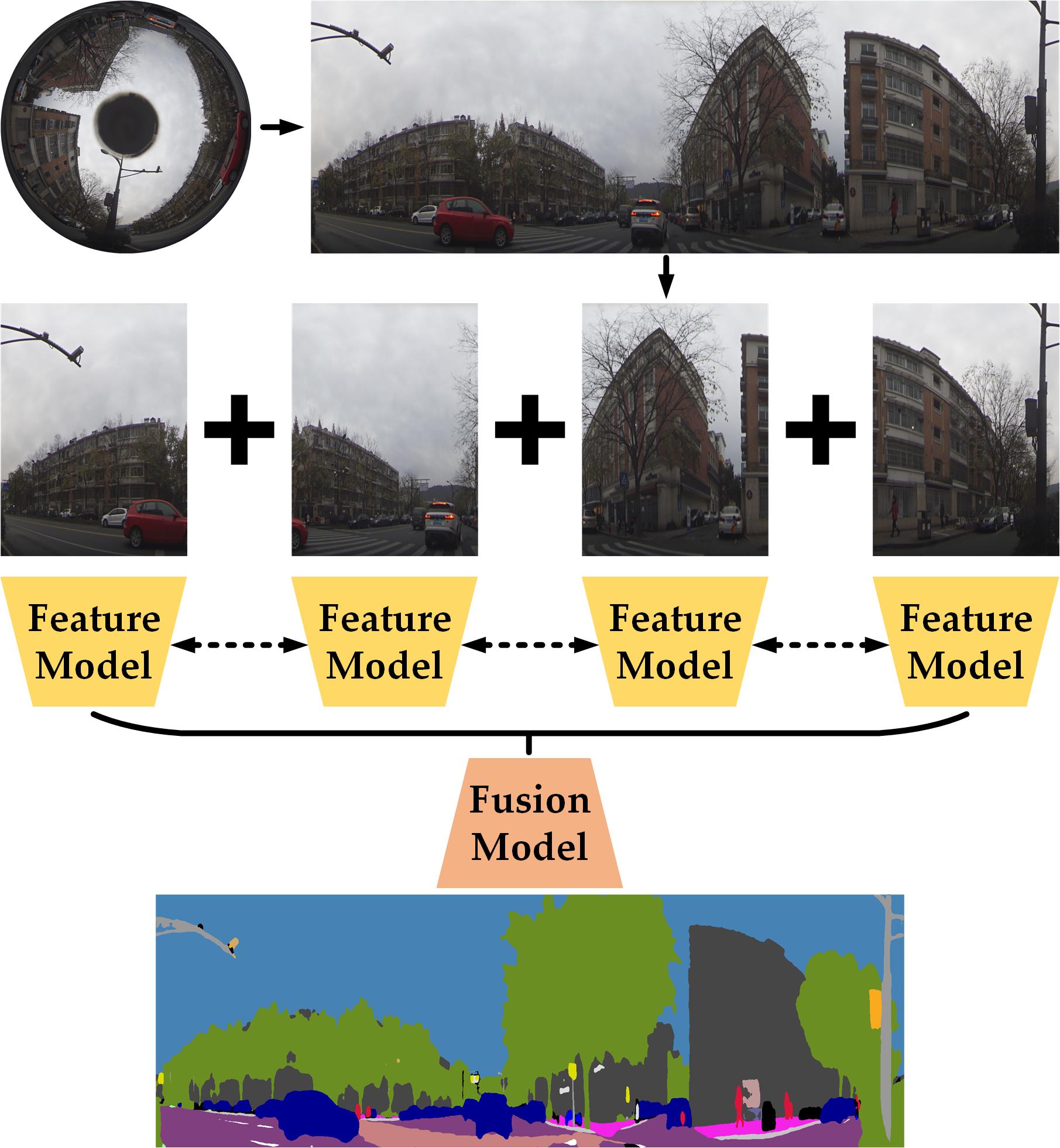}
\caption{The proposed framework for panoramic annular semantic segmentation. Each feature model (corresponding to the single feature model like encoder in conventional architectures) is responsible for predicting the semantically-meaningful high-level feature map of a panorama segment while interacting with the neighboring ones through cross-segment padding (indicated by the dotted arrows). Fusion model incorporates the feature maps and completes the panoramic segmentation.}
\label{figurelabel_framework}
\vskip-4ex
\end{figure}

Fig.~\ref{figurelabel_framework} depicts the diagram of the proposed network adaptation framework. Efficient semantic segmentation networks generally pursue the encoder-decoder fashion~\cite{badrinarayanan2017segnet}. In this work, for a given semantic segmentation network, it is methodologically re-separated into a feature model and a fusion model, to facilitate the adaptation from pinhole source domain to panoramic target domain. The feature model could be equivalent to the encoder (which encodes semantically-meaningful features), while the fusion model corresponds to the decoder (which fuses the features), or the re-separation position could vary with the target to realize continuous panoramic segmentation kept in mind.

The panoramic annular image is firstly unfolded using the interface offered by the toolbox for omnidirectional camera calibration~\cite{scaramuzza2006toolbox}. After that, the unfolded panorama is partitioned into several segments (4 segments illustrated in Fig.~\ref{figurelabel_framework}). Each panorama segment will be fed into the feature model of a semantic segmentation network to obtain semantically-meaningful dense feature maps.
This adaptation is due to that although a large-scale panoramic segmentation dataset is not available for training, there is a correspondence between the features inferred from each panorama segment and the features learned from pinhole image dataset. In this regard, robust ultra-wide FoV panoramic segmentation is reachable by leveraging conventional narrow-FoV training data that has already demonstrated diversity in terms of geographical location and observational viewpoint~\cite{neuhold2017mapillary}. Finally, the feature maps of different segments will be concatenated and max-pooled (along the unfolding direction) to be fed into the fusion model to yield the final panoramic pixel-wise classification results. This is due to that the fusion model with lean convolutional layers is mainly responsible for the classification when the semantically-meaningful feature map has already been predicted and aggregated. The fusion model has to be no leaner to enable continuous segmentation as it could take a wide context around the borders of panorama segments into consideration.

In addition, to prevent blind spots around the borders which is critical in surrounding scene perception, each feature model interacts with the neighboring one (see Fig.~\ref{figurelabel_framework}) to enable real 360$^{\circ}$ processing. More precisely, we extend ring-padding~\cite{payen2018eliminating} that is also termed as circular convolution~\cite{schubert2019circular}, by replacing traditional zero-padding around the panorama border with the specialized padding by copying features from the opposite side of the unfolded panorama, which theoretically eliminates the inherent image border. Our key extension lies in the specialized padding near the border of panorama segments where padding is taken from the neighboring feature map, without creating blind spots in the interchanges of the FoVs (segments), such that surrounding environments can be seamlessly interpreted.

\subsection{SwaftNet: Proposed Network with Attention Connections}

\begin{figure}[t]
\setlength{\abovecaptionskip}{0pt}
\setlength{\belowcaptionskip}{0pt}
\centering
\includegraphics[width=0.48\textwidth]{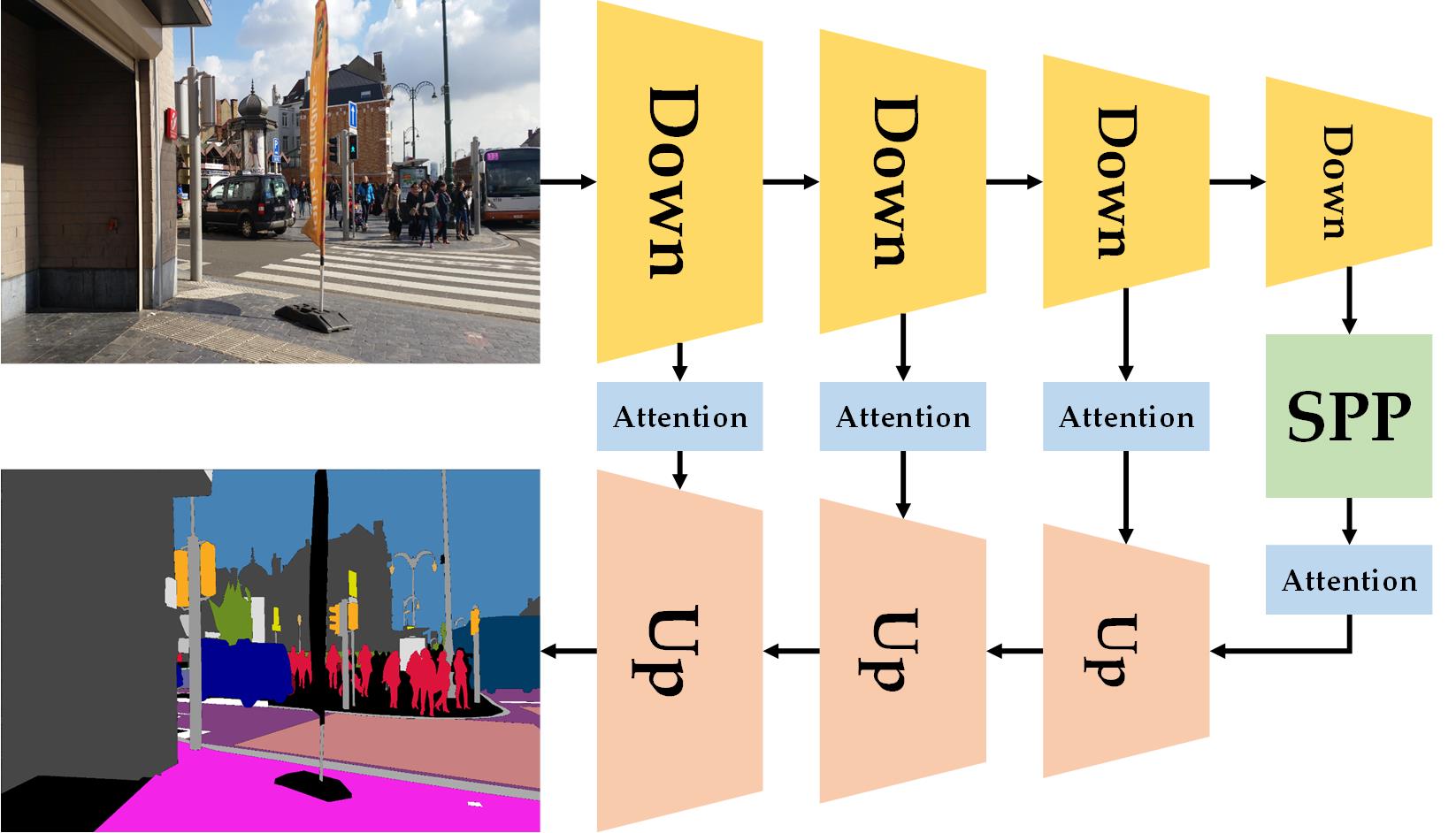}
\caption{The proposed architecture with attention-based lateral connections to blend semantically-rich deep layers with spatially-detailed shallow layers. The down-sampling path with the SPP module (encoder) corresponds to the feature model in Fig.~\ref{figurelabel_framework}, while the up-sampling path (decoder) corresponds to the fusion model.}
\label{figurelabel_swiftnet}
\vskip-4ex
\end{figure}

To enhance the detail-sensitivity of semantic segmentation which is especially critical for panoramic images, we spotlight the low-level spatially-rich features exploited at the down-sampling stages by using lateral attentional connections as shown in Fig.~\ref{figurelabel_swiftnet}. The proposed architecture follows the single-scale model of SwiftNet~\cite{orsic2019defense} that transforms the input RGB image into dense pixel-level semantic predictions, throughout a down-sampling encoder and up-sampling decoder, which is based on an U-shape structure like U-Net~\cite{ronneberger2015u} and LinkNet~\cite{chaurasia2017linknet}. Specifically, the encoder and the decoder are asymmetric, as the dimensionality of encoder features increases along the down-sampling path, while the dimensionality of the decoder features keeps constant. This is because recognition (encoder) requires more capacity when rough semantic content is known. We therefore opt for a simple decoder composed of minimalistic up-sampling modules. Accordingly, it is necessary to adjust the dimensionality to blend features from both paths, e.g., by using 1$\times$1 convolutions in the lateral connections~\cite{orsic2019defense}.

On the other hand, we argue that the informative spatially-detailed features could be further spotlighted. With the purpose of increasing sensitivity to useful components and suppressing less informative features, we replace the 1$\times$1 convolutions with attention operations~\cite{hu2018squeeze}, by squeezing spatial information into a channel descriptor, and element-wisely multiplying the feature map taken from the corresponding down-sampling stage and the re-shaped descriptor. The re-weighted feature map has the same dimensionality as the feature map at the corresponding up-sampling stage. The feature maps from both paths will be added, resulting an attention-augmented lateral connection. The key to enhance detail-sensitivity lies in that the conventional 1$\times$1 convolution-based lateral connections are adapted to attention operations as the skip-connections in our architecture. The proposed SwaftNet is 
further adapted by leveraging our network adaptation framework in a general way. In addition, we insert a channel-wise attention module after the Spatial Pyramid Pooling (SPP) module~\cite{orsic2019defense} that is a simplified version of pyramidal pooling module from PSPNet~\cite{zhao2017pyramid}. In our architecture, SPP features must pass through the entire decoder (up-sampling path) before getting classified. Hence, it acts as an instrument to enlarge the receptive field. The channel attention (squeeze-excite operation) is light-weight~\cite{hu2018squeeze}, which enables adaptive feature refinement, hence the re-calibrated detail-sensitive feature responses have been further highlighted.

\section{Experiments}

\subsection{Datasets and Setups}

\begin{figure}[t]
\setlength{\abovecaptionskip}{0pt}
\setlength{\belowcaptionskip}{0pt}
\centering
\includegraphics[width=0.48\textwidth]{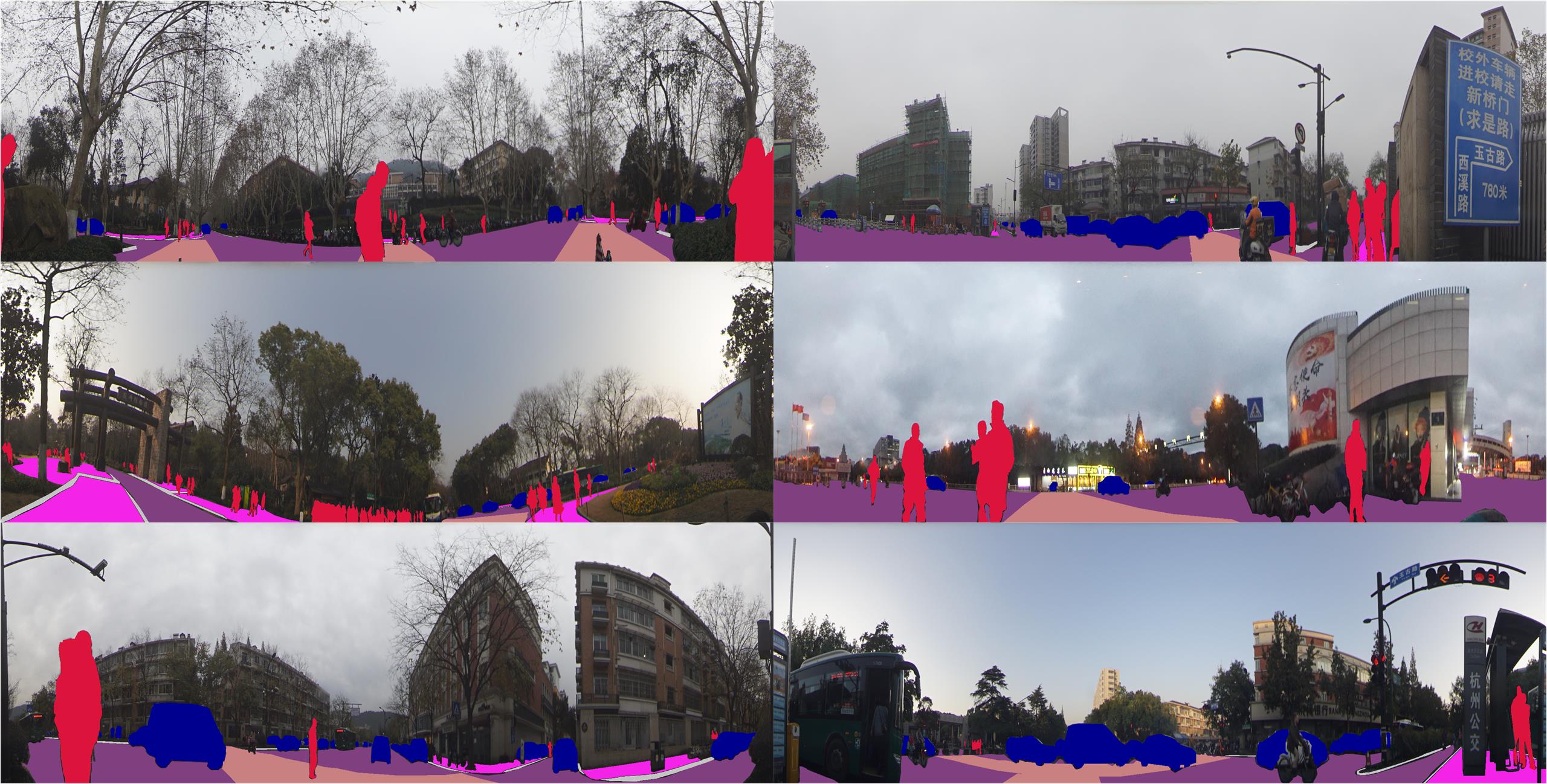}
\caption{Examples from the extended PASS dataset with annotations.}
\label{figurelabel_dataset}
\vskip-4ex
\end{figure}

For evaluation purposes, the Mapillary Vistas validation dataset (2000 samples)~\cite{neuhold2017mapillary} and the Panoramic Annular Semantic Segmentation (PASS) dataset (400 images)~\cite{yang2019can} are used. The original PASS dataset contains 4 classes, precisely Road, Car, Crosswalk and Curb which are important for large-scale road scene understanding. We further manually annotate the 400 images on two classes: Sidewalk and Person, whose segmentation are more detail sensitivity-desired, yet exceptionally safety-critical. The panoramic images have a 2048$\times$692 (1.4 megapixel) resolution. Example panoramas for testing with pixel-fine annotations are shown in Fig.~\ref{figurelabel_dataset}.

Mapillary Vistas training set (18000 images) is chosen to train our panoramic scene parsers, leveraging its high diversity in focal length, viewpoint, as well as geographical location which are both important for robust segmentation. We take the critical 27 classes as listed in Table~\ref{table_accuracy} to adapt to our campus scenarios. We apply a heterogeneous set of data augmentation operations~\cite{yang2019can}, separating in geometry, texture, distortion and style transformations, which are of special relevance to the generalizability, when a trained semantic segmentation model is required to work reliably in new, unseen domains.
Our architecture is trained under Adam optimization, starting with a learning rate of 1.0$\times$10$^{-4}$ that decreases exponentially until reaching convergence with a batch size of 10 and an input resolution of 1024$\times$512.

\begin{table*}[t]
\scriptsize
\setlength{\abovecaptionskip}{0pt}
\setlength{\belowcaptionskip}{0pt}
\caption{Segmentation accuracy of the proposed SwaftNet on Mapillary Vistas dataset.}
\vskip-2ex
\label{table_accuracy}
\begin{center}

\begin{tabular}{|m{2.5em}<{\centering}|m{2.5em}<{\centering}|m{2.5em}<{\centering}|m{2.5em}<{\centering}|m{2.5em}<{\centering}|m{2.5em}<{\centering}|m{2.5em}<{\centering}|m{2.5em}<{\centering}|m{2.5em}<{\centering}|m{2.5em}<{\centering}|m{2.5em}<{\centering}|m{2.5em}<{\centering}|m{2.5em}<{\centering}|m{2.5em}<{\centering}|}
\hline
{\textbf{\rotatebox{90}{Pole}}}&{\textbf{\rotatebox{90}{Street Light}}}&{\textbf{\rotatebox{90}{Billboard}}}&{\textbf{\rotatebox{90}{Traffic Light } }}&{\textbf{\rotatebox{90}{Car}}}&{\textbf{\rotatebox{90}{Truck}}}&{\textbf{\rotatebox{90}{Bicycle}}}&{\textbf{\rotatebox{90}{Motorcycle}}}&{\textbf{\rotatebox{90}{Bus}}}&{\textbf{\rotatebox{90}{Sign Front}}}&{\textbf{\rotatebox{90}{Sign Back}}}&{\textbf{\rotatebox{90}{Road}}}&{\textbf{\rotatebox{90}{Sidewalk}}}&{\textbf{\rotatebox{90}{Curb Cut}}}\\
\hline
{47.5\%}&{35.8\%}&{43.4\%}&{62.8\%}&{90.3\%}&{70.4\%}&{55.9\%}&{59.1\%}&{75.1\%}&{69.5\%}&{38.7\%}&{88.6\%}&{68.8\%}&{14.7\%}\\
\hline
{\textbf{\rotatebox{90}{Plain}}}&{\textbf{\rotatebox{90}{Bike Lane }}}&{\textbf{\rotatebox{90}{Curb}}}&{\textbf{\rotatebox{90}{Fence}}}&{\textbf{\rotatebox{90}{Wall}}}&{\textbf{\rotatebox{90}{Building}}}&{\textbf{\rotatebox{90}{Person}}}&{\textbf{\rotatebox{90}{Rider}}}&{\textbf{\rotatebox{90}{Sky}}}&{\textbf{\rotatebox{90}{Vegetation }}}&{\textbf{\rotatebox{90}{Terrain} }}&{\textbf{\rotatebox{90}{Marking}}}&{\textbf{\rotatebox{90}{Crosswalk }}}&{\textbf{mIoU}}\\
\hline
{17.4\%}&{37.3\%}&{55.5\%}&{55.0\%}&{46.7\%}&{86.6\%}&{69.9\%}&{47.3\%}&{98.2\%}&{89.7\%}&{63.7\%}&{53.5\%}&{62.3\%}&{59.4\%}\\
\hline
\end{tabular}
\end{center}
\vskip-4ex
\end{table*}

In this way, the proposed SwaftNet achieves a mean Intersection over Union (mIoU) of 59.4\% on the Vistas validation set (2000 images) as shown in Table~\ref{table_accuracy}, which demonstrates the learning capacity against large dataset. For the classes corresponding to the evaluation classes of PASS dataset, including Road (88.6\%), Car (90.3\%), Sidewalk (68.8\%), Crosswalk (62.3\%), Curb (55.5\%) and Person (69.9\%), accuracies are readily qualified, such that we step further to study on the robustness aspect by assessing the performance in the panoramic imagery domain.

\subsection{Benchmarking Panoramic Semantic Segmentation}

With the extended PASS dataset, we benchmark the performance of state-of-the-art networks especially those designed for efficient semantic segmentation as real-time setup is crucial for autonomous transportation applications, including SegNet~\cite{badrinarayanan2017segnet}, ENet~\cite{paszke2016enet}, LinkNet~\cite{chaurasia2017linknet}, EDANet~\cite{lo2018efficient}, BiSeNet~\cite{yu2018bisenet}, CGNet~\cite{wu2018cgnet}, ERFNet~\cite{romera2019bridging} and PSPNet18~\cite{zhao2017pyramid}. We take the proposed training hyper-parameters for them in their respective publications, but all with same data augmentation techniques to facilitate fair comparison.
We also contrast with the baseline method ERF-PSPNet that was preliminarily used for the task Panoramic Annular Semantic Segmentation~\cite{yang2019can}, where the pyramid pooling module has been similarly employed with channel attention operations to form as a variant named ERF-APSPNet~\cite{yang2019pass}. The per-class results on PASS dataset are shown in Table~\ref{table_comparison}, where Fig.~\ref{figurelabel_acc} provides a visualization of the comparison.

\begin{table*}[h]
\scriptsize
\setlength{\abovecaptionskip}{0pt}
\setlength{\belowcaptionskip}{0pt}
\caption{Comparison of state-of-the-art efficient networks on Mapillary VISTAS validation dataset and PASS testing dataset.}
\vskip-2ex
\label{table_comparison}
\begin{center}

\begin{tabular}{|c|c|c|c|c|c|c|c|c|c|}
\hline
\multirow{2}*{\textbf{Network}}&{\textbf{On VISTAS}}&\multicolumn{8}{c|}{\textbf{On PASS}}\\
\cline{2-10}
&{\textbf{mIoU}}&{\textbf{Car}}&{\textbf{Road}}&{\textbf{Sidewalk}}&{\textbf{Crosswalk}}&{\textbf{Curb}}&{\textbf{Person}}&{\textbf{mIoU}}&{\textbf{mIoU boost}}\\
\hline
{SegNet~\cite{badrinarayanan2017segnet}}&{48.4\%}&{82.3\%}&{72.0\%}&{31.0\%}&{32.7\%}&{18.1\%}&{48.0\%}&{47.4\%}&{18.0\%}\\
\hline
{ENet~\cite{paszke2016enet}}&{47.0\%}&{89.2\%}&{76.2\%}&{43.4\%}&{53.5\%}&{29.0\%}&{60.2\%}&{58.6\%}&{27.8\%}\\
\hline
{LinkNet~\cite{chaurasia2017linknet}}&{47.7\%}&{87.4\%}&{77.4\%}&{42.9\%}&{47.4\%}&{22.1\%}&{47.5\%}&{54.1\%}&{23.9\%}\\
\hline
{EDANet~\cite{lo2018efficient}}&{49.6\%}&{85.0\%}&{78.3\%}&{40.6\%}&{51.5\%}&{21.7\%}&{49.0\%}&{54.4\%}&{24.1\%}\\
\hline
{BiSeNet~\cite{yu2018bisenet}}&{49.5\%}&{83.1\%}&{65.0\%}&{41.9\%}&{18.1\%}&{13.5\%}&{35.0\%}&{42.8\%}&{15.3\%}\\
\hline
{CGNet~\cite{wu2018cgnet}}&{52.8\%}&{72.8\%}&{73.6\%}&{29.5\%}&{37.9\%}&{15.1\%}&{18.9\%}&{41.3\%}&{11.2\%}\\
\hline
{ERFNet~\cite{romera2019bridging}}&{52.7\%}&{91.0\%}&{76.1\%}&{49.0\%}&{56.4\%}&{27.9\%}&{59.6\%}&{60.0\%}&{25.8\%}\\
\hline
{PSPNet18~\cite{zhao2017pyramid}}&{50.4\%}&{89.0\%}&{77.1\%}&{47.2\%}&{46.9\%}&{26.9\%}&{42.9\%}&{55.0\%}&{20.4\%}\\
\hline
{ERF-PSPNet~\cite{yang2018unifying}\cite{yang2019can}}&{52.1\%}&{91.4\%}&{77.6\%}&{43.5\%}&{52.9\%}&{33.4\%}&{67.2\%}&{61.0\%}&{22.0\%}\\
\hline
{ERF-APSPNet}&{54.8\%}&{91.0\%}&{78.6\%}&{46.8\%}&{57.2\%}&{30.8\%}&{72.8\%}&{62.9\%}&{27.6\%}\\
\hline
{SwiftNet~\cite{orsic2019defense}}&{52.4\%}&{91.3\%}&{\textbf{79.7\%}}&{43.6\%}&{55.9\%}&{34.6\%}&{72.2\%}&{62.9\%}&{25.7\%}\\
\hline
{SwaftNet}&{\textbf{59.4\%}}&{\textbf{93.6\%}}&{77.6\%}&{\textbf{53.7\%}}&{\textbf{62.1\%}}&{\textbf{38.3\%}}&{\textbf{80.7\%}}&{\textbf{67.7\%}}&{\textbf{29.9\%}}\\
\hline
\end{tabular}

\end{center}
\vskip-4ex
\end{table*}

\begin{figure}[t]
\setlength{\abovecaptionskip}{0pt}
\setlength{\belowcaptionskip}{0pt}
\centering
\includegraphics[width=0.48\textwidth]{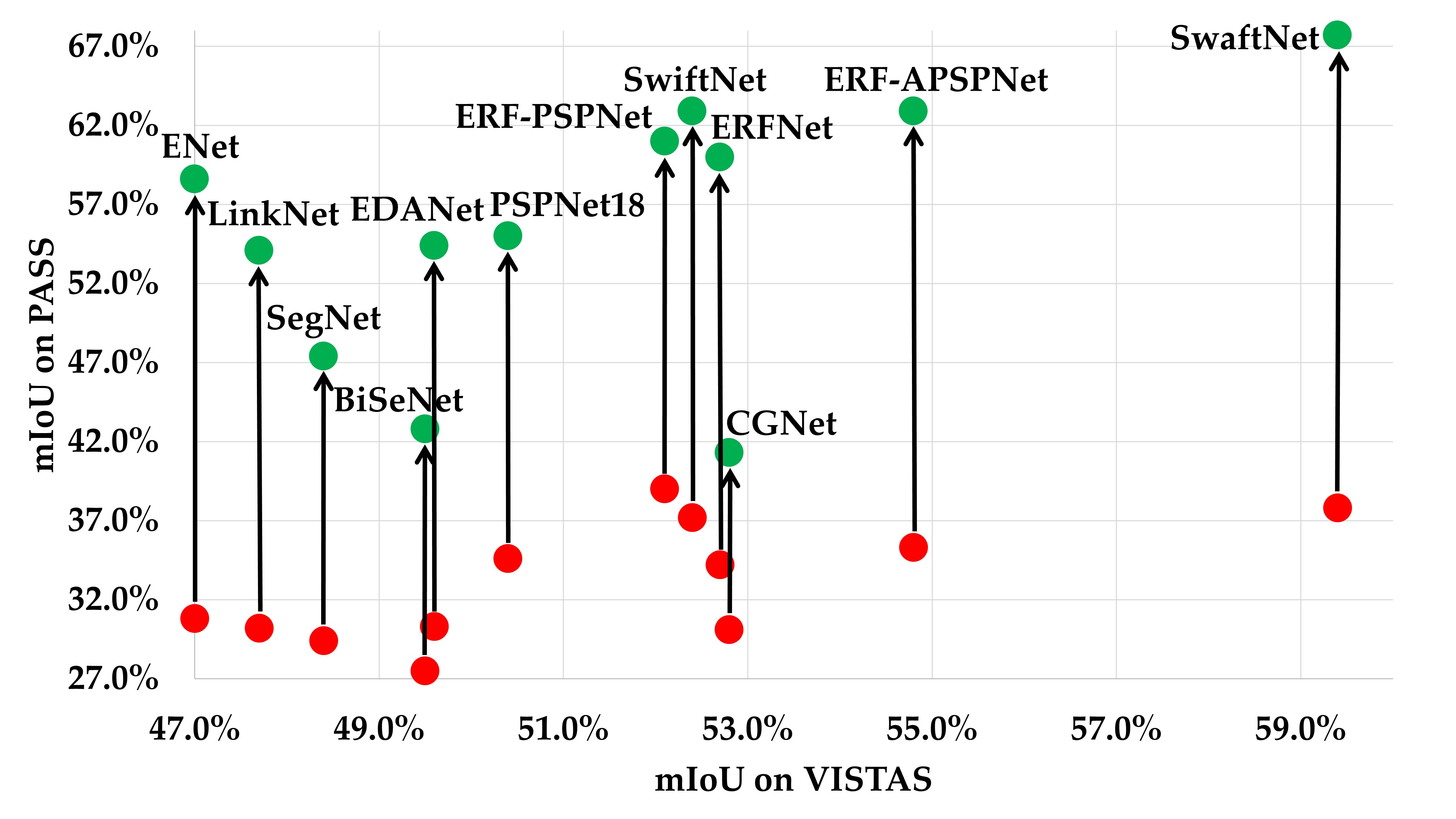}
\caption{Segmentation accuracy in mIoU across Mapillary VISTAS validation and PASS testing datasets. Red points: accuracies of networks without adaptation; Green points: with adaptation, which is consistently effective for state-of-the-art efficient networks.}
\label{figurelabel_acc}
\vskip-4ex
\end{figure}

It turns out the proposed network SwaftNet outperforms the known efficient networks measured in mIoU on Mapillary Vistas validation set and the extended PASS dataset. Compared with ERF-PSPNet and its variant ERF-APSPNet which achieved state-of-the-art results on the PASS dataset~\cite{yang2019can}, the proposed model outperforms them by large margins, precisely 6.7\% and 4.8\% respectively. The proposed usage of attention lateral connections enables to exploit more informative low-level features, such that it leads to mIoU increases of 7.0\% and 4.8\% over the basic SwiftNet on the respective dataset. Our network slightly falls behind on Road which is usually close to the dark textures near the panorama outer borders, so it becomes more cautious to classify those areas as safe roadways. More importantly, on other safety-critical yet detail-sensitive classes, remarkable improvements have been observed: Sidewalk (10.1\%), Crosswalk (6.2\%), Curb (3.7\%) and Person (8.5\%). 

In addition to the new state-of-the-art scores yielded by our SwaftNet, we find that the proposed adaptation strategy is consistently and significantly positive for all the trained networks, as illustrated in Fig.~\ref{figurelabel_acc}. This is verified by comparing the performance of the networks in two settings: deploying with or without the adaptation framework, where encoder is used as feature model, and fusion model corresponds to decoder. We take the conclusion in~\cite{yang2019pass} which finds that when a single model is tested with a FoV of 90$^{\circ}$, it achieves the best performance. Accordingly, for an adapted network in 360$^{\circ}$ imagery, it is deployed with 4 features models (just as Fig.~\ref{figurelabel_framework} shows), each of which interacts with the neighboring one. In the other setting, each network is tested in an end-to-end way by viewing the whole panorama as a single segment without any 360$^{\circ}$ adaptation. As shown in Table~\ref{table_comparison}, for ENet, ERFNet and SwiftNet, they have even attained IoU boosts of over 25.0\%, demonstrating the importance of network adaptation when taken to the panoramic imagery. The results also show that the adaptation is not strictly tied with a specific architecture, but can be effective and deployed with different segmentation models. Even more, the proposed network achieves 29.9\% of mIoU boost. When the segment number is appropriate, details exploited by the feature model are also more adapted to leverage the detail-aware capability featured by our proposed network, such that the benefit of the network adaptation is more remarkable than being deployed with other architectures.

\begin{figure*}[h]
\setlength{\abovecaptionskip}{0pt}
\setlength{\belowcaptionskip}{0pt}
\centering
\includegraphics[width=1.0\textwidth]{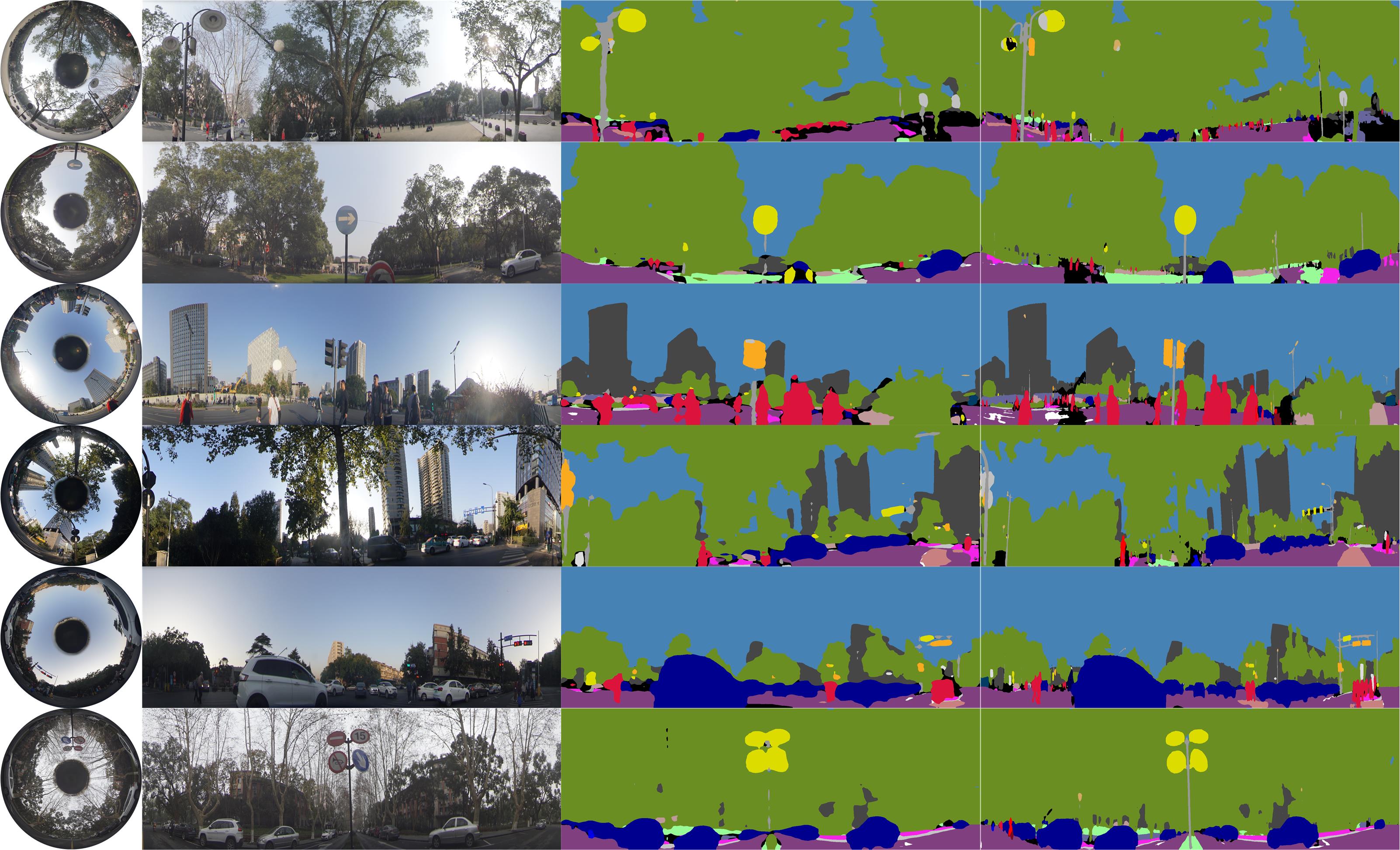}
\captionsetup[subfloat]{captionskip=-10pt}
\subfloat[\label{figurelabel_qualitative_pass_a}Raw]{\hspace{0.1\textwidth}}
\subfloat[\label{figurelabel_qualitative_pass_b}Unfolded panorama]{\hspace{0.3\textwidth}}
\subfloat[\label{figurelabel_qualitative_pass_c}Baseline (ERF-PSPNet)]{\hspace{0.3\textwidth}}
\subfloat[\label{figurelabel_qualitative_pass_d}Ours (SwaftNet)]{\hspace{0.3\textwidth}}
\vskip-1ex
\caption{Qualitative examples of panoramic annular semantic segmentation: (a) Raw annular images from the PASS dataset, (b) Unfolded panoramas, (c)~Segmentation maps yielded by ERF-PSPNet (coarse, detail lost) and (d) SwaftNet (fine, detail preserved).}
\label{figurelabel_qualitative_pass}
\vskip-3ex
\end{figure*}

Moreover, Fig.~\ref{figurelabel_qualitative_pass} displays the segmentation maps of example panoramas from the PASS dataset, which were parsed by using the baseline method ERF-PSPNet and our proposed SwaftNet. It can be easily seen that our approach enables finer segmentation, accurately preserving the details of many small classes such as the pedestrians and traffic signs, as well as the tiny traffic scene elements like the thin poles, traffic lights and the cars in the distance. In these senses, panoramic annular semantic segmentation has been endowed with detail-sensitivity evidenced by both numerical and qualitative parsing results.

\subsection{Real-World Performance with Views to Real Deployment}

\begin{figure}[t]
\setlength{\abovecaptionskip}{0pt}
\setlength{\belowcaptionskip}{0pt}
\centering
\includegraphics[width=0.48\textwidth]{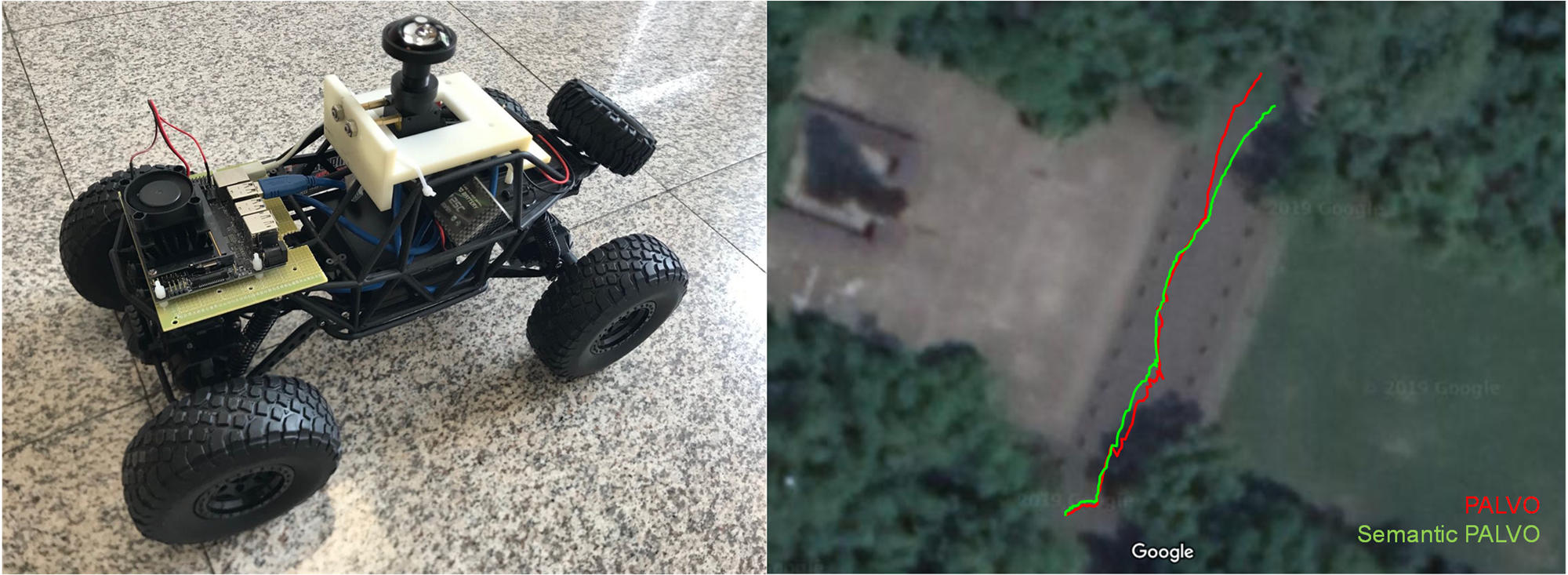}
\captionsetup[subfloat]{captionskip=-10pt}
\subfloat[\label{figurelabel_vo_a}Mobile robot]{\hspace{0.24\textwidth}}
\subfloat[\label{figurelabel_vo_b}Reconstructed trajectories]{\hspace{0.24\textwidth}}
\caption{(a) The mobile robot installed with Panoramic Annular Lens and NVIDIA Jetson Nano; (b) Reconstructed trajectories by a state-the-of-art visual odometry PALVO with/without predicted panoramic semantics.}
\label{figurelabel_vo}
\vskip-4ex
\end{figure}

\begin{figure*}[h]
\setlength{\abovecaptionskip}{0pt}
\setlength{\belowcaptionskip}{0pt}
\centering
\includegraphics[width=1.0\textwidth]{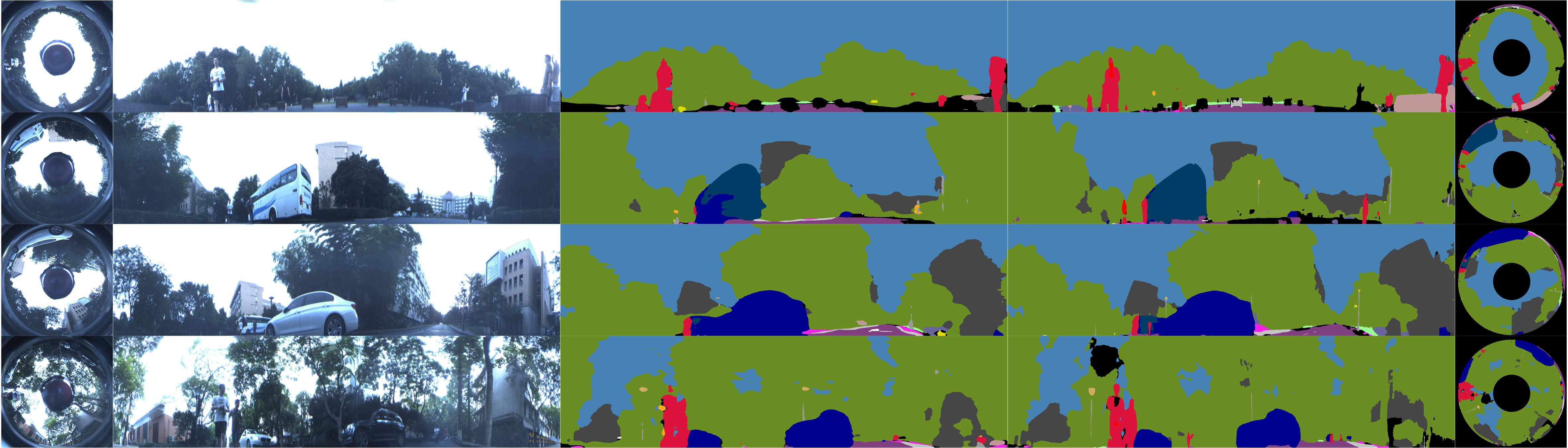}
\captionsetup[subfloat]{captionskip=-10pt}
\subfloat[\label{figurelabel_qualitative_ugv_a}Raw]{\hspace{0.071\textwidth}}
\subfloat[\label{figurelabel_qualitative_ugv_b}Unfolded panorama]{\hspace{0.286\textwidth}}
\subfloat[\label{figurelabel_qualitative_ugv_c}Baseline (ERF-PSPNet)]{\hspace{0.286\textwidth}}
\subfloat[\label{figurelabel_qualitative_ugv_d}Ours (SwaftNet)]{\hspace{0.286\textwidth}}
\subfloat[\label{figurelabel_qualitative_ugv_e}PASS]{\hspace{0.071\textwidth}}
\vskip-1ex
\caption{(a) Raw annular images captured by Panoramic Annular Lens installed on the mobile robot, (b) Unfolded panoramas, (c) Segmentation maps yielded by ERF-PSPNet and (d) SwaftNet, (e) Folded-back PASS maps. Raw annular images and folded-back semantic maps are fed to PALVO~\cite{chen2019palvo}.}
\label{figurelabel_qualitative_ugv}
\end{figure*}

\begin{figure*}[t]
\setlength{\abovecaptionskip}{0pt}
\setlength{\belowcaptionskip}{0pt}
\centering
\includegraphics[width=1.0\textwidth]{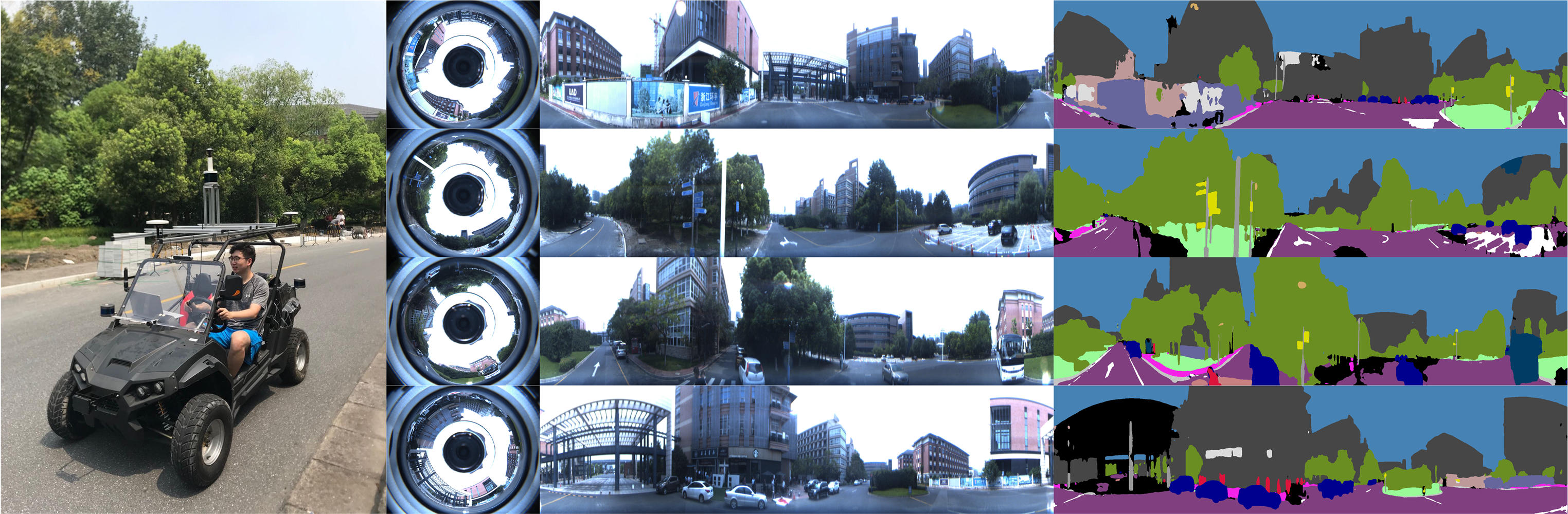}
\captionsetup[subfloat]{captionskip=-10pt}
\subfloat[\label{figurelabel_instrumentation_a}Instrumented vehicle]{\hspace{0.246\textwidth}}
\subfloat[\label{figurelabel_instrumentation_b}Raw]{\hspace{0.098\textwidth}}
\subfloat[\label{figurelabel_instrumentation_c}Unfolded panorama]{\hspace{0.328\textwidth}}
\subfloat[\label{figurelabel_instrumentation_d}DS-PASS]{\hspace{0.328\textwidth}}
\vskip-1ex
\caption{(a) The instrumented electric vehicle installed with Panoramic Annular Lens and LiDAR sensors, (b) Raw annular images, (c) Unfolded panoramas, (d) Panoramic segmentation maps produced by our DS-PASS system with suitable FoVs for intelligent vehicles applications.}
\label{figurelabel_instrumentation}
\vskip-3ex
\end{figure*}

Stepping further than validating on the PASS dataset which was originally captured by a wearable personal navigation assistance prototype~\cite{yang2019can}, the proposed framework has been deployed on a mobile robot (see Fig.~\ref{figurelabel_vo_a} and Fig.~\ref{figurelabel_qualitative_ugv_d}) and an instrumented vehicle (see Fig.~\ref{figurelabel_instrumentation}). With a view to real deployment in autonomous transportation applications like these two systems, real-time setup and real-world robustness are two of the most important factors that should be taken into consideration. 

To measure the inference speed, we test the adapted models running through the 400 panoramas, where each feature model is fed with a 1024$\times$512 segment, as shown in Table~\ref{table_cityscapes}. We also include the mIoU results when training on Cityscapes dataset~\cite{cordts2016cityscapes} and evaluating on its validation set using 19 classes. We use the implementation of \cite{xiang2019comparative} to facilitate the comparison on Cityscapes.
The proposed network reaches 88.9FPS on a Titan RTX GPU processor, which is slightly slower than the basic SwiftNet (96.5FPS), but the proposed usage of attention skip-connection helps to attain huge accuracy boosts on diverse datasets, especially on VISTAS which contains more small and detail-critical classes than Cityscapes. Compared with the state-of-the-art method ERF-PSPNet~\cite{yang2019can} (40.2FPS) for the PASS task and its variant ERF-APSPNet (38.0FPS), SwaftNet is significantly faster. Overall, our proposed SwaftNet holds an excellent trade-off between accuracy and efficiency, while for panoramic annular semantic segmentation, it is not only more efficient but also more precise than the baseline work.

\begin{table}[h]
\scriptsize
\setlength{\abovecaptionskip}{0pt}
\setlength{\belowcaptionskip}{0pt}
\caption{Accuracy (in mIoU) and speed (in FPS) analysis.}
\vskip-2ex
\label{table_cityscapes}
\begin{center}

\begin{tabular}{|c||c||c|c|c|}
\hline
{\textbf{Network}}&{\textbf{Cityscapes}}&{\textbf{VISTAS}}&{\textbf{PASS (FPS)}}\\
\hline
{ERFNet~\cite{romera2019bridging}}&{65.8\%}&{52.7\%}&{60.0\% (34.7)}\\
\hline
{ERF-PSPNet~\cite{yang2018unifying}\cite{yang2019can}}&{64.1\%}&{52.1\%}&{61.0\% (40.2)}\\
\hline
{SwiftNet~\cite{orsic2019defense}\cite{xiang2019comparative}}&{69.8\%}&{52.4\%}&{62.9\% (\textbf{96.5)}}\\
\hline
{Our SwaftNet}&{\textbf{72.1\%}}&{\textbf{59.4\%}}&{\textbf{67.7\%} (88.9)}\\
\hline
\end{tabular}

\end{center}
\vskip-4ex
\end{table}

We use a NVIDIA Jetson Nano that is very portable to be deployed on the mobile robot, where a single forward pass of our network achieves near real-time computation (22.15FPS at the resolution of 640$\times$480). Fig.~\ref{figurelabel_qualitative_ugv} shows representative segmentation maps for the captured panoramas when the mobile robot is navigating through the campus. Since the panoramic annular lens system is installed at a lower position (see Fig.~\ref{figurelabel_vo_a}), the perspective is significantly different to that of the training samples, as a large part of images in Mapillary Vistas are captured from vehicles. In spite of the significantly lower viewpoint, panoramic segmentation based on our SwaftNet maintains highly qualified and robust as shown in Fig.~\ref{figurelabel_qualitative_ugv_d}, meanwhile more details are well preserved than the baseline ERF-PSPNet (Fig.~\ref{figurelabel_qualitative_ugv_c}).

For the instrumented vehicle, we use a newly designed panoramic annular lens system, which images a FoV of 360$^{\circ}\times$70$^{\circ}$. When installing the camera on top of the instrumented electric vehicle as shown in Fig.~\ref{figurelabel_instrumentation_a}, it has a FoV of 40$^{\circ}$ above the horizontal plane, and a surrounding view with 30$^{\circ}$ vertical FoV below, which allows to perceive more information about the roadways than the mobile robot. 
As the FoV is very matched with the instrumented vehicle, the observed roadways are beneficial for sensing the street structure, which leads to clear and well-defined semantic maps as displayed in Fig.~\ref{figurelabel_instrumentation_d}, making our DS-PASS system ideally suitable for IV applications.

\subsection{Semantic Panoramic Visual Odometry}

\begin{figure}[t]
\setlength{\abovecaptionskip}{0pt}
\setlength{\belowcaptionskip}{0pt}
\centering
\includegraphics[width=0.48\textwidth]{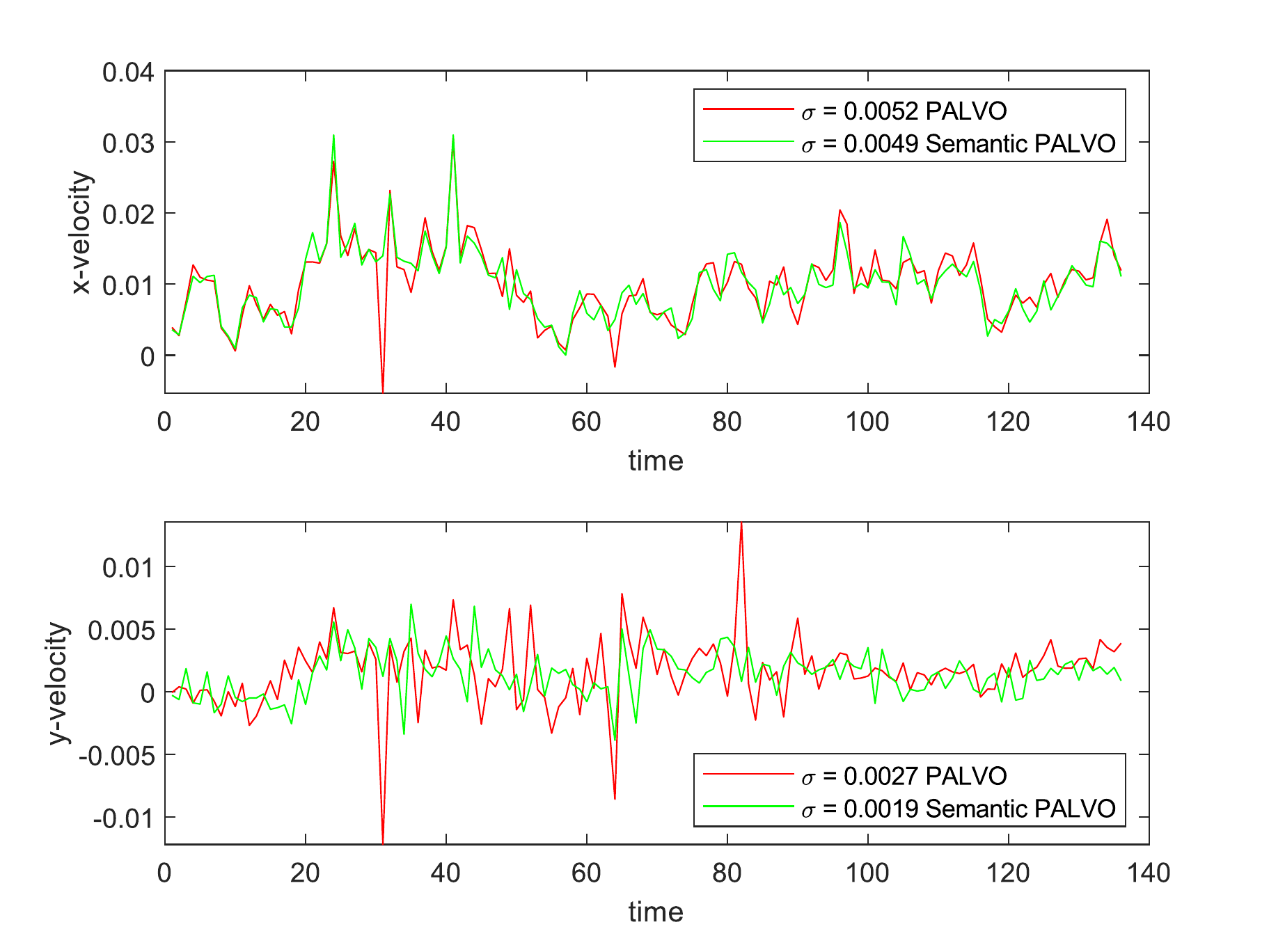}
\caption{Velocity curves produced by the panoramic visual odometry with/without using semantics predicted by our DS-PASS system. Note that the odometry based on PALVO~\cite{chen2019palvo} is scale-invariant.}
\label{figurelabel_velocity}
\vskip-4ex
\end{figure}

In order to investigate the benefit of the predicted panoramic semantics for upper-level navigational tasks, we leverage a state-of-the-art visual odometry algorithm PALVO~\cite{chen2019palvo} that has achieved excellent robustness against rapid motions and dynamic scene elements. Based on PALVO, we build a semantic panoramic visual odometry system. First, the SwaftNet-predicted 360$^{\circ}$ semantics for the unfolded panoramas (Fig.~\ref{figurelabel_qualitative_ugv_d}) are projected/folded back to the raw annular image as shown in Fig.~\ref{figurelabel_qualitative_ugv_e}. Second, the semantic maps are fed to the odometry jointly with the raw panoramas (Fig.~\ref{figurelabel_qualitative_ugv_a}). 

More specifically, the semantic information are used to determine the key points. Only the matched points with consistent semantic labels across frames are leveraged to compute the 6-DoF pose, otherwise they will be regarded as outliers which will not be involved for pose estimation. As shown in Fig.~\ref{figurelabel_vo_b}, finally the trajectory (more than 50m) produced by the semantic PALVO system is smoother than that of the original PALVO. As the mobile robot moves at a steady speed, the velocity would not change drastically. This is supported by the velocity curves visualized in Fig.~\ref{figurelabel_velocity}, as the standard deviations of velocity estimated by semantic PALVO are smaller in both directions.

Overall, the higher-level robustness of the semantic panoramic visual odometry system is attributed to two factors: the large FoV to robustify against swift motion and dynamic real-world scenarios; and the panoramic semantics to reduce the drift and smooth the trajectory. Moreover, we find that the semantic PALVO system working with the baseline network frequently leads to failure, which is due to the less accurate and detail-lost semantic maps, such that the inlier points are often too sparse to accurately estimate the pose and successfully track across frames. In this sense, the superiority of the proposed DS-PASS system with SwaftNet is further validated by deploying with the state-of-the-art PALVO algorithm. 

\section{CONCLUSIONS}

In this paper, we propose a Detail-Sensitive Panoramic Annular Semantic Segmentation (DS-PASS) framework to perceive the surroundings.
DS-PASS is presented with a general network adaptation method that enables state-of-the-art semantic segmentation network trained on perspective imaging datasets to work comfortably in panoramic imagery domain. Specifically, we put forward a real-time semantic segmentation network named SwaftNet, by blending semantically-meaningful deep layers with spatially-detailed shallow layers using lateral attentional skip-connections, to perform segmentation both swiftly and accurately. 

The network adaptation method is demonstrated to be consistently effective, allowing state-of-the-art networks to re-use the knowledge learned from pinhole images, and parse annular images in a suitable and comfortable way. This unlocks the usage of panoramic cameras for countless semantic sensing-desired systems, bypassing the laborious and time-consuming training label acquisition. Our proposed SwaftNet yields the new state-of-the-art scores on the extended challenging PASS dataset with more annotated classes to reflect detail-sensitivity. Furthermore, the proposed DS-PASS framework has been deployed on a mobile robot and an instrumented vehicle, demonstrating the real-time speed and the real-world robustness.

With the mobile robot, we show that when the vertical perspective is not ideal, the segmentation maintains highly qualified, and the predicted panoramic semantics are beneficial for upstream visual odometry algorithms like the state-of-the-art PALVO. With the instrumented fully electric vehicle, we show that the proposed DS-PASS system is ideally tailored for Intelligent Vehicles (IV) applications, which only needs a single camera to address surrounding sensing in a single pass, leaving rich opportunities to be fused with LiDAR point clouds.



\bibliographystyle{IEEEtran}
\bibliography{bib.bib}

\begin{thebibliography}{10}
\providecommand{\url}[1]{#1}
\csname url@samestyle\endcsname
\providecommand{\newblock}{\relax}
\providecommand{\bibinfo}[2]{#2}
\providecommand{\BIBentrySTDinterwordspacing}{\spaceskip=0pt\relax}
\providecommand{\BIBentryALTinterwordstretchfactor}{4}
\providecommand{\BIBentryALTinterwordspacing}{\spaceskip=\fontdimen2\font plus
\BIBentryALTinterwordstretchfactor\fontdimen3\font minus
  \fontdimen4\font\relax}
\providecommand{\BIBforeignlanguage}[2]{{%
\expandafter\ifx\csname l@#1\endcsname\relax
\typeout{** WARNING: IEEEtran.bst: No hyphenation pattern has been}%
\typeout{** loaded for the language `#1'. Using the pattern for}%
\typeout{** the default language instead.}%
\else
\language=\csname l@#1\endcsname
\fi
#2}}
\providecommand{\BIBdecl}{\relax}
\BIBdecl

\bibitem{yang2018unifying}
K.~Yang, L.~M. Bergasa, E.~Romera, R.~Cheng, T.~Chen, and K.~Wang, ``Unifying
  terrain awareness through real-time semantic segmentation,'' in \emph{2018
  IEEE Intelligent Vehicles Symposium (IV)}.\hskip 1em plus 0.5em minus
  0.4em\relax IEEE, 2018, pp. 1033--1038.

\bibitem{cordts2016cityscapes}
M.~Cordts, M.~Omran, S.~Ramos, T.~Rehfeld, M.~Enzweiler, R.~Benenson,
  U.~Franke, S.~Roth, and B.~Schiele, ``The cityscapes dataset for semantic
  urban scene understanding,'' in \emph{2016 IEEE Conference on Computer Vision
  and Pattern Recognition (CVPR)}.\hskip 1em plus 0.5em minus 0.4em\relax IEEE,
  2016, pp. 3213--3223.

\bibitem{neuhold2017mapillary}
G.~Neuhold, T.~Ollmann, S.~R. Bul{\`o}, and P.~Kontschieder, ``The mapillary
  vistas dataset for semantic understanding of street scenes,'' in \emph{2017
  IEEE International Conference on Computer Vision (ICCV)}.\hskip 1em plus
  0.5em minus 0.4em\relax IEEE, 2017, pp. 5000--5009.

\bibitem{yang2019can}
K.~Yang, X.~Hu, L.~M. Bergasa, E.~Romera, X.~Huang, D.~Sun, and K.~Wang, ``Can
  we pass beyond the field of view? panoramic annular semantic segmentation for
  real-world surrounding perception,'' in \emph{2019 IEEE Intelligent Vehicles
  Symposium (IV)}.\hskip 1em plus 0.5em minus 0.4em\relax IEEE, 2019, pp.
  446--453.

\bibitem{paszke2016enet}
A.~Paszke, A.~Chaurasia, S.~Kim, and E.~Culurciello, ``Enet: A deep neural
  network architecture for real-time semantic segmentation,'' \emph{arXiv
  preprint arXiv:1606.02147}, 2016.

\bibitem{romera2019bridging}
E.~Romera, L.~M. Bergasa, K.~Yang, J.~M. Alvarez, and R.~Barea, ``Bridging the
  day and night domain gap for semantic segmentation,'' in \emph{2019 IEEE
  Intelligent Vehicles Symposium (IV). IEEE}, 2019, pp. 1312--1318.

\bibitem{yang2019pass}
K.~Yang, X.~Hu, L.~M. Bergasa, E.~Romera, and K.~Wang, ``Pass: Panoramic
  annular semantic segmentation,'' \emph{IEEE Transactions on Intelligent
  Transportation Systems}, 2019.

\bibitem{chen2019palvo}
H.~Chen, K.~Wang, W.~Hu, K.~Yang, R.~Cheng, X.~Huang, and J.~Bai, ``Palvo:
  visual odometry based on panoramic annular lens,'' \emph{Optics express},
  vol.~27, no.~17, pp. 24\,481--24\,497, 2019.

\bibitem{long2015fully}
J.~Long, E.~Shelhamer, and T.~Darrell, ``Fully convolutional networks for
  semantic segmentation,'' in \emph{2015 IEEE Conference on Computer Vision and
  Pattern Recognition (CVPR)}.\hskip 1em plus 0.5em minus 0.4em\relax IEEE,
  2015, pp. 3431--3440.

\bibitem{ronneberger2015u}
O.~Ronneberger, P.~Fischer, and T.~Brox, ``U-net: Convolutional networks for
  biomedical image segmentation,'' in \emph{International Conference on Medical
  image computing and computer-assisted intervention}.\hskip 1em plus 0.5em
  minus 0.4em\relax Springer, 2015, pp. 234--241.

\bibitem{badrinarayanan2017segnet}
V.~Badrinarayanan, A.~Kendall, and R.~Cipolla, ``Segnet: A deep convolutional
  encoder-decoder architecture for image segmentation,'' \emph{IEEE
  transactions on pattern analysis and machine intelligence}, vol.~39, no.~12,
  pp. 2481--2495, 2017.

\bibitem{zhao2017pyramid}
H.~Zhao, J.~Shi, X.~Qi, X.~Wang, and J.~Jia, ``Pyramid scene parsing network,''
  in \emph{2017 IEEE Conference on Computer Vision and Pattern Recognition
  (CVPR)}.\hskip 1em plus 0.5em minus 0.4em\relax IEEE, 2017, pp. 6230--6239.

\bibitem{hu2019acnet}
X.~Hu, K.~Yang, L.~Fei, and K.~Wang, ``Acnet: Attention based network to
  exploit complementary features for rgbd semantic segmentation,'' in
  \emph{2019 IEEE International Conference on Image Processing (ICIP)}.\hskip
  1em plus 0.5em minus 0.4em\relax IEEE, 2019, pp. 1440--1444.

\bibitem{chaurasia2017linknet}
A.~Chaurasia and E.~Culurciello, ``Linknet: Exploiting encoder representations
  for efficient semantic segmentation,'' in \emph{2017 IEEE Visual
  Communications and Image Processing (VCIP)}.\hskip 1em plus 0.5em minus
  0.4em\relax IEEE, 2017, pp. 1--4.

\bibitem{orsic2019defense}
M.~Orsic, I.~Kreso, P.~Bevandic, and S.~Segvic, ``In defense of pre-trained
  imagenet architectures for real-time semantic segmentation of road-driving
  images,'' in \emph{Proceedings of the IEEE Conference on Computer Vision and
  Pattern Recognition}, 2019, pp. 12\,607--12\,616.

\bibitem{xiang2019comparative}
K.~Xiang, K.~Wang, and K.~Yang, ``A comparative study of high-recall real-time
  semantic segmentation based on swift factorized network,'' \emph{arXiv
  preprint arXiv:1907.11394}, 2019.

\bibitem{deng2017cnn}
L.~Deng, M.~Yang, Y.~Qian, C.~Wang, and B.~Wang, ``Cnn based semantic
  segmentation for urban traffic scenes using fisheye camera,'' in \emph{2017
  IEEE Intelligent Vehicles Symposium (IV)}.\hskip 1em plus 0.5em minus
  0.4em\relax IEEE, 2017, pp. 231--236.

\bibitem{deng2019restricted}
L.~Deng, M.~Yang, H.~Li, T.~Li, B.~Hu, and C.~Wang, ``Restricted deformable
  convolution-based road scene semantic segmentation using surround view
  cameras,'' \emph{IEEE Transactions on Intelligent Transportation Systems},
  2019.

\bibitem{yahiaoui2019fisheyemodnet}
M.~Yahiaoui, H.~Rashed, L.~Mariotti, G.~Sistu, I.~Clancy, L.~Yahiaoui, V.~R.
  Kumar, and S.~Yogamani, ``Fisheyemodnet: Moving object detection on
  surround-view cameras for autonomous driving,'' \emph{arXiv preprint
  arXiv:1908.11789}, 2019.

\bibitem{iordache2019low}
L.~Iordache, V.~Paunescu, W.~Kang, J.~Kwon, A.~Leica, and B.~Jeon,
  ``Low-complexity scene understanding network,'' in \emph{International
  Conference on Image Analysis and Processing}.\hskip 1em plus 0.5em minus
  0.4em\relax Springer, 2019, pp. 245--256.

\bibitem{sakurada2018weakly}
K.~Sakurada, ``Weakly supervised silhouette-based semantic change detection,''
  \emph{arXiv preprint arXiv:1811.11985}, 2018.

\bibitem{pan2019cross}
B.~Pan, J.~Sun, A.~Andonian, A.~Oliva, and B.~Zhou, ``Cross-view semantic
  segmentation for sensing surroundings,'' \emph{arXiv preprint
  arXiv:1906.03560}, 2019.

\bibitem{zhang2019orientation}
C.~Zhang, S.~Liwicki, W.~Smith, and R.~Cipolla, ``Orientation-aware semantic
  segmentation on icosahedron spheres,'' \emph{arXiv preprint
  arXiv:1907.12849}, 2019.

\bibitem{jiang2019dfnet}
W.~Jiang, Y.~Wu, L.~Guan, and J.~Zhao, ``Dfnet: Semantic segmentation on
  panoramic images with dynamic loss weights and residual fusion block,'' in
  \emph{2019 International Conference on Robotics and Automation (ICRA)}.\hskip
  1em plus 0.5em minus 0.4em\relax IEEE, 2019, pp. 5887--5892.

\bibitem{romera2018erfnet}
E.~Romera, J.~M. Alvarez, L.~M. Bergasa, and R.~Arroyo, ``Erfnet: Efficient
  residual factorized convnet for real-time semantic segmentation,'' \emph{IEEE
  Transactions on Intelligent Transportation Systems}, vol.~19, no.~1, pp.
  263--272, 2018.

\bibitem{scaramuzza2006toolbox}
D.~Scaramuzza, A.~Martinelli, and R.~Siegwart, ``A toolbox for easily
  calibrating omnidirectional cameras,'' in \emph{2006 IEEE/RSJ International
  Conference on Intelligent Robots and Systems}.\hskip 1em plus 0.5em minus
  0.4em\relax IEEE, 2006, pp. 5695--5701.

\bibitem{payen2018eliminating}
G.~Payen~de La~Garanderie, A.~Atapour~Abarghouei, and T.~P. Breckon,
  ``Eliminating the blind spot: Adapting 3d object detection and monocular
  depth estimation to 360 panoramic imagery,'' in \emph{Proceedings of the
  European Conference on Computer Vision (ECCV)}, 2018, pp. 789--807.

\bibitem{schubert2019circular}
S.~Schubert, P.~Neubert, J.~P{\"o}schmann, and P.~Protzel, ``Circular
  convolutional neural networks for panoramic images and laser data,'' in
  \emph{2019 IEEE Intelligent Vehicles Symposium (IV)}.\hskip 1em plus 0.5em
  minus 0.4em\relax IEEE, 2019, pp. 653--660.

\bibitem{hu2018squeeze}
J.~Hu, L.~Shen, and G.~Sun, ``Squeeze-and-excitation networks,'' in \emph{2018
  IEEE/CVF Conference on Computer Vision and Pattern Recognition}.\hskip 1em
  plus 0.5em minus 0.4em\relax IEEE, 2018, pp. 7132--7141.

\bibitem{lo2018efficient}
S.-Y. Lo, H.-M. Hang, S.-W. Chan, and J.-J. Lin, ``Efficient dense modules of
  asymmetric convolution for real-time semantic segmentation,'' \emph{arXiv
  preprint arXiv:1809.06323}, 2018.

\bibitem{yu2018bisenet}
C.~Yu, J.~Wang, C.~Peng, C.~Gao, G.~Yu, and N.~Sang, ``Bisenet: Bilateral
  segmentation network for real-time semantic segmentation,'' in
  \emph{Proceedings of the European Conference on Computer Vision (ECCV)},
  2018, pp. 325--341.

\bibitem{wu2018cgnet}
T.~Wu, S.~Tang, R.~Zhang, and Y.~Zhang, ``Cgnet: A light-weight context guided
  network for semantic segmentation,'' \emph{arXiv preprint arXiv:1811.08201},
  2018.

\end{thebibliography}

\end{document}